\documentclass[conference]{IEEEtran}
\IEEEoverridecommandlockouts

\usepackage{cite}
\usepackage{xurl}
\usepackage{amsmath,amssymb,amsfonts}
\usepackage{algorithm}
\usepackage{algorithmic}
\usepackage{graphicx}
\usepackage{textcomp}
\usepackage{xcolor}
\usepackage{booktabs}
\usepackage{multirow}
\usepackage{tikz}
\usepackage{pgfplots}
\pgfplotsset{compat=1.18}
\usepackage{pgfplots}
\usepackage{comment}
\pgfplotsset{compat=1.18}
\usepgfplotslibrary{groupplots}
\usetikzlibrary{arrows.meta, positioning, shapes.geometric, calc}

\usepgfplotslibrary{groupplots,fillbetween}
\usetikzlibrary{
    arrows.meta,
    positioning,
    shapes.geometric,
    calc,
    intersections
}

\def\BibTeX{{\rm B\kern-.05em{\sc i\kern-.025em b}\kern-.08em
    T\kern-.1667em\lower.7ex\hbox{E}\kern-.125emX}}

\newcommand{\Design}[0]{\textsc{LIMBO}}

\begin{document}

\title{LIMBO: Lifelong Inference-Time Memory and Budget Optimization for LLM Agents}


\author{
    \IEEEauthorblockN{
        Siddharth Sharma\textsuperscript{1},
        Nilesh Prasad Pandey\textsuperscript{1},
        Onat Gungor\textsuperscript{2},
        Tajana Rosing\textsuperscript{1}
    }
    \IEEEauthorblockA{
        \textsuperscript{1}University of California, San Diego, CA, USA \\
        \textsuperscript{2}West Virginia University, Morgantown, WV, USA \\
        \{sis004, nppandey, tajana\}@ucsd.edu, onat.gungor@wvu.edu
    }
}



\maketitle

\begin{abstract}
As LLM agents become integrated into increasingly complex workflows, they must continually acquire new capabilities while retaining competence on previously learned tasks. Lifelong agents address this through experience replay, injecting past interactions into the prompt to leverage prior experience during inference. However, replay is not free: every replayed trajectory competes with retrieval, reasoning, tool use, and verification for the same limited prompt and compute budget, making effective resource allocation essential. Existing approaches allocate these resources using fixed replay policies, regardless of whether replay is beneficial for the current task. We identify this as inference-time memory allocation, a distinct problem class for lifelong agents, and introduce \Design{}: the first online framework to our knowledge that treats memory as a controllable inference-time resource and jointly optimizes memory strategy and inference budget for each incoming task. Unlike prior approaches that fix the replay policy or require model
weights, teacher supervision, or offline retraining, \Design{} learns this allocation online in a single pass, explicitly balancing task performance and inference cost without modifying the underlying agent. Across three LLM backbones on LifelongAgentBench, \Design{} achieves
better cost-accuracy tradeoffs than state-of-the-art memory-augmented baselines, and nearly matches all strongest such baselines at up to $\sim$83\% lower inference cost ($\sim$53\% on
average).  \Design{} adapts its policy across models and environments without retraining, demonstrating that effective allocation can be learned online rather than manually specified.

\end{abstract}


\section{Introduction}
\label{sec:introduction}

LLM agents are increasingly deployed in environments where they must solve a continuous stream of evolving tasks while accumulating experience over time~\cite{liu2023agentbench, zhou2024webarena, koh2024visualwebarena}. Unlike conventional single-task settings, these agents can leverage accumulated experiences from previous interactions to improve future performance through mechanisms such as memory, retrieval, and tool use~\cite{jiang2024longllmlingua, yu2025adaptive_reasoning_ibpo, lyu2025hbpo,wen2025budgetthinker, li2025knapsack_rl}. However, these mechanisms compete for limited inference-time resources, including context length, computational budget, and latency, creating a fundamental tradeoff between leveraging accumulated experience and maintaining inference efficiency~\cite{efficientagents2025}. As lifelong agents encounter diverse tasks and environments, effectively allocating these resources becomes critical for preserving prior knowledge while adapting to new challenges~\cite{zheng2025lab}. 
\begin{figure}[t]
\centering
\begin{tikzpicture}
\begin{axis}[
    width=0.95\columnwidth,
    height=0.65\columnwidth,
    xlabel={Cost per task (\$)},
    ylabel={Accuracy (\%)},
    xmin=0,
    xmax=0.32,
    ymin=70,
    ymax=77,
    grid=major,
    grid style={gray!15},
    tick label style={font=\footnotesize},
    label style={font=\footnotesize},
    title style={font=\footnotesize\bfseries},
    clip=false,
]

\addplot[
    thick,
    dashed,
    gray!70,
    mark=none
]
coordinates {
    (0.0459,73.0)
    (0.167,74.5)
    (0.2896,74.85)
};

\addplot[
    only marks,
    mark=*,
    mark size=2.8pt,
    color=gray!70!black
]
coordinates {
    (0.0459,73.0)
};

\node[
    font=\scriptsize,
    anchor=north
]
at (axis cs:0.0459,72.8)
{No replay};

\addplot[
    only marks,
    mark=star,
    mark size=4.5pt,
    color=violet!70!black
]
coordinates {
    (0.0504,74.4)
};

\node[
    font=\scriptsize,
    text=violet!70!black,
    anchor=south
]
at (axis cs:0.0504,74.7)
{\Design{}};

\node[
    font=\scriptsize\itshape,
    text=violet!70!black,
    align=center
]
at (axis cs:0.075,71.4)
{1.10$\times$ cost\\+1.4 pts};

\addplot[
    only marks,
    mark=*,
    mark size=2.8pt,
    color=blue!65!black
]
coordinates {
    (0.167,74.5)
};

\node[
    font=\scriptsize,
    anchor=south
]
at (axis cs:0.167,74.8)
{$k=8$};

\addplot[
    only marks,
    mark=*,
    mark size=2.8pt,
    color=blue!65!black
]
coordinates {
    (0.2896,74.85)
};

\node[
    font=\scriptsize,
    anchor=south
]
at (axis cs:0.2896,75.15)
{$k=16$};

\node[
    font=\scriptsize\itshape,
    text=gray!60!black,
    align=center
]
at (axis cs:0.120,73.15)
{3.6$\times$ cost\\+1.5 pts};

\node[
    font=\scriptsize\itshape,
    text=gray!60!black,
    align=center
]
at (axis cs:0.245,73.15)
{1.7$\times$ cost\\+0.35 pts};

\end{axis}
\end{tikzpicture}

\caption{Replay cost--accuracy tradeoff on DBBench. Relative to no replay,
\Design{} improves accuracy by 1.4 points while increasing cost by only
1.10$\times$ (9.8\%). In contrast, increasing the replay budget from no
replay to $k=8$ raises cost by 3.6$\times$ for a 1.5-point gain, while
increasing it further from $k=8$ to $k=16$ incurs another 1.7$\times$
cost increase for only 0.35 additional points.}
\label{fig:cost-utility-mismatch}
\end{figure}
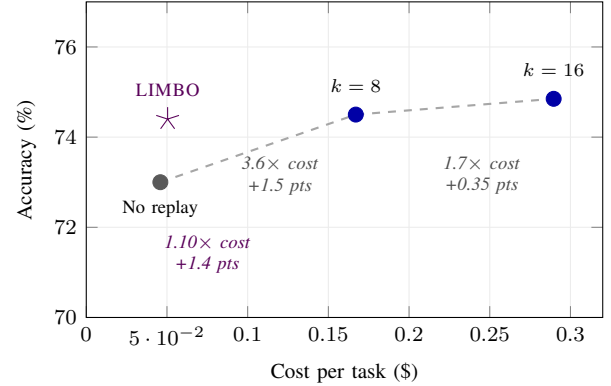

Memory represents one of the most critical components in this resource-allocation problem, as lifelong agents must decide which past experiences to retain and retrieve under limited inference-time budgets. Among existing approaches, \emph{experience replay} has become a widely adopted mechanism for leveraging historical interactions~\cite{zheng2025lab}. Experience replay injects past trajectories into the current task prompt, enabling agents to reuse previously acquired tool-use patterns, command templates, and action-format conventions without requiring parameter updates. 
However, despite its effectiveness in improving agent performance, replay introduces significant scalability challenges as agents accumulate more experience. As the task count $t$ increases, the replay buffer grows proportionally, resulting in an $\mathcal{O}(t)$ increase in the number of injected tokens and corresponding inference-time overhead. Moreover, the utility of individual trajectories can vary substantially depending on the current task, model, and environment. Consequently, uniform replay strategies may incur considerable context and computation costs by allocating resources to trajectories that provide limited benefit for the current task.

This cost-value mismatch is not merely a theoretical concern; it appears empirically across environments and replay strategies. Figure~\ref{fig:cost-utility-mismatch} illustrates that increasing fixed replay buffers provides diminishing performance improvements while incurring rapidly growing inference costs. Scaling fixed replay
from no-replay to $k{=}16$ on DBBench~\cite{zheng2025lab} with Qwen2.5 7B raises inference cost by more than $5\times$ but improves accuracy by only  2 points, with the $k{=}8 \to k{=}16$ step being effectively flat. Compression-based approaches~\cite{jiang2024longllmlingua, jiang2023llmlingua} reduce token overhead but tend to disrupt the structured action traces required by interactive environments. Retrieval-based strategies~\cite{isele2018selective} improve memory relevance but still introduce additional selection costs and consume limited context capacity. These limitations reveal a fundamental challenge: effective lifelong agents must determine not only which experiences to replay, but also when replay provides sufficient value to justify its inference-time cost. More broadly, agents must dynamically allocate limited resources across memory, retrieval, reasoning, tool use, and verification, which all compete for the same context and computation budgets.

We address this challenge with \Design{}, a lifelong controller that adaptively allocates inference-time memory and compute resources. For each incoming task, \Design{} selects a memory strategy and compute budget using a lightweight contextual bandit based on prompt and runtime features. This formulation provides a unified online allocation framework over existing memory mechanisms, including fixed replay, prompt compression, selective replay~\cite{isele2018selective, hazard2025sure}, and memory-augmented agents~\cite{zhang2026memrl, jaglan2025atlas}. Across three LLM backbones evaluated on LifelongAgentBench, \Design{} achieves comparable or superior performance to the strongest fixed replay baseline while reducing inference cost by up to $\sim$83\%. Furthermore, \Design{} learns different memory allocation policies across model and environment configurations, highlighting the importance of adaptive inference-time resource management.

\section{Related Work}
\label{sec:related}

\textbf{Lifelong Learning and LLM Agent Benchmarks.} Lifelong learning aims to retain knowledge across sequential tasks
without forgetting~\cite{french1999catastrophic}, with classical
approaches based on regularization~\cite{kirkpatrick2017overcoming} and experience replay~\cite{isele2018selective}. These methods target static supervised settings; extending them to LLM agents that act and
adapt across interactive environments remains largely
unexplored~\cite{zheng2025lab}, and prior replay methods carry no mechanism for
adaptive inference time budget allocation.  Existing agent benchmarks such as
WebArena~\cite{zhou2024webarena}, AgentBench~\cite{liu2023agentbench}, and
VisualWebArena~\cite{koh2024visualwebarena} evaluate isolated tasks in static
environments, with no protocol for sequential skill transfer.
LifelongAgentBench (LAB)~\cite{zheng2025lab} addresses this gap with three
interactive environments, explicit skill labels, a skill
dependency graph, and automatic label verification, providing a clean setting
in which fixed, compressed, retrieved, and gated replay can be compared under
one protocol.

\textbf{Agent Efficiency.} Recent work on adaptive reasoning and budget-aware inference
studies how much computation an LLM should spend at test
time~\cite{yu2025adaptive_reasoning_ibpo,lyu2025hbpo,wen2025budgetthinker,li2025knapsack_rl,wei2022cot},
and cascade methods learn budget constrained routing across models or inference
paths~\cite{zhang2024efficient_contextual_cascade}.  Empirical analysis of
agent cost confirms that efficiency is task dependent and poorly addressed by
static design choices~\cite{efficientagents2025}.  In the lifelong setting
specifically, LAB's group self consistency~\cite{zheng2025lab} stabilizes
performance under large replay buffers but at higher token cost than na\"ive
replay, moving against the efficiency objective.  \Design{} brings inference
time compute control to the lifelong setting and treats memory use itself as a controllable action.

\textbf{Replay Representation and Selection.}
Recent work has explored replay from several perspectives, including prompt compression, memory selection, and structured memory representations. LongLLMLingua~\cite{jiang2024longllmlingua,jiang2023llmlingua} compresses prompts to reduce token usage, but replay trajectories are sequential action demonstrations rather than static documents, making compression alone insufficient. Isele and Cosgun~\cite{isele2018selective} proposed principled replay selection criteria, while SuRe~\cite{hazard2025sure} prioritizes replay using surprise scores computed from dual LoRA adapters, requiring access to model weights and limiting applicability to black-box LLMs. MemRL~\cite{zhang2026memrl}, ATLAS~\cite{jaglan2025atlas}, CER~\cite{liu2025cer}, and JitRL~\cite{li2026jitrl} improve memory representation or retrieval, but do not explicitly optimize the inference-time cost of replay under a unified lifelong learning framework. In contrast, \Design{} formulates replay as a cost-aware inference-time decision problem, selecting replay actions on a per-sample basis to balance performance and computational cost.

\section{\Design{} Framework}
\subsection{Overview}

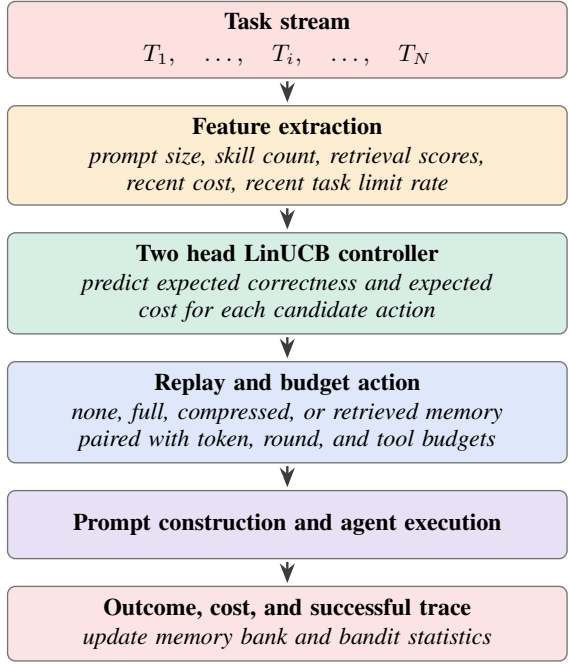
\begin{figure}[t]
\centering
\definecolor{cInput}{RGB}{253,224,224}
\definecolor{cFeat}{RGB}{253,237,210}
\definecolor{cCtrl}{RGB}{215,238,228}
\definecolor{cAct}{RGB}{222,232,250}
\definecolor{cExec}{RGB}{232,225,245}
\definecolor{cOut}{RGB}{252,228,228}
\definecolor{cEdge}{RGB}{60,60,60}
\begin{tikzpicture}[
    font=\small,
    node distance=3.5mm,
    stage/.style={
        rectangle, rounded corners=3pt,
        draw=black!55, line width=0.5pt,
        text width=0.80\columnwidth,
        inner sep=4.5pt, align=center,
        minimum height=9mm
    },
    arr/.style={
        ->, >={Stealth[length=2.6mm,width=2mm]},
        line width=0.9pt, draw=cEdge, shorten >=1pt, shorten <=1pt
    },
    label/.style={font=\scriptsize\itshape, midway, right, xshift=2pt}
]
\node[stage, fill=cInput] (s1)
  {\textbf{Task stream}\\[2pt]
   $T_1,\;\ldots,\;T_i,\;\ldots,\;T_N$};

\node[stage, fill=cFeat, below=of s1] (s2)
  {\textbf{Feature extraction}\\[1pt]
   \textit{prompt size, skill count, retrieval scores,}\\
   \textit{recent cost, recent task limit rate}};

\node[stage, fill=cCtrl, below=of s2] (s3)
  {\textbf{Two head LinUCB controller}\\[1pt]
   \textit{predict expected correctness and expected}\\
   \textit{cost for each candidate action}};

\node[stage, fill=cAct, below=of s3] (s4)
  {\textbf{Replay and budget action}\\[1pt]
   \textit{none, full, compressed, or retrieved memory}\\
   \textit{paired with token, round, and tool budgets}};

\node[stage, fill=cExec, below=of s4] (s5)
  {\textbf{Prompt construction and agent execution}};

\node[stage, fill=cOut, below=of s5] (s6)
  {\textbf{Outcome, cost, and successful trace}\\[1pt]
   \textit{update memory bank and bandit statistics}};

\draw[arr] (s1) -- (s2);
\draw[arr] (s2) -- (s3);
\draw[arr] (s3) -- (s4);
\draw[arr] (s4) -- (s5);
\draw[arr] (s5) -- (s6);
\end{tikzpicture}
\caption{\Design{}: Online memory allocation loop. The controller selects a replay action and compute budget before each task, then updates the memory bank and bandit statistics using the observed reward and cost after execution.}
\label{fig:framework}
\end{figure}

Figure~\ref{fig:framework} presents our \Design{} instantiation of lifelong
inference allocation, which includes:
(i) a stream of interactive tasks, (ii) a memory bank of successful interaction
trajectories, (iii) online feature extraction from the current prompt, the
retrieval state, and recent runtime statistics, (iv) a two head contextual
bandit that predicts correctness and cost for each candidate action, and
(v) prompt construction and agent execution under the selected replay mode
and compute budget.  After each task completes, the observed outcome and
measured cost update both the memory bank and the bandit statistics.

\Design{} turns replay from a fixed preprocessing step into an online compute
allocation decision.  The memory bank stores only completed trajectories with
measured outcomes and cost.  For a new task, the controller compares memory
modes and compute budgets using features that are available before the agent
acts.  The chosen action determines how much history enters the prompt and how
much execution budget the agent receives.  After the agent runs once, observed
correctness and cost update both the memory bank and the controller.

\subsection{Problem Setup}

Let $\mathcal{T} = (T_1, T_2, \ldots, T_N)$ be a sequential task stream.  At
step $i$, the agent receives task $T_i$, executes an interactive session, and
produces a final answer.  After completion, the benchmark returns a binary
correctness signal $y_i \in \{0,1\}$ and a measured cost $c_i \ge 0$.  We
write the completed session as $\sigma_i = (T_i, y_i, c_i, \tau_i)$, where
$\tau_i$ is the full interaction trajectory: the task prompt, the agent's
actions at each round, the environment observations and tool outputs they
produced, and the final answer. The agent maintains a memory buffer of previously successful sessions,
\[
\mathcal{M}_i = \{\sigma_j : j < i,\; y_j = 1\}\,,
\]
which can be injected into the prompt for future tasks.  Whether injection helps depends on relevance: a past trajectory transfers value primarily when it shares operations, schema, or action format with the current task.
For example, a past DBBench session executing a \texttt{JOIN} with \texttt{GROUP BY} is informative for a new task requiring the same SQL primitives, but adds only prompt overhead for an unrelated single-table \texttt{SELECT}.

The measured cost $c_i$ is the total token count of the completed
session (prompt plus generation) multiplied by the provider's
per-token schedule, or the standardized schedule described in
Section~\ref{sec:setup} for local backbones.  Because absolute
cost magnitudes vary by orders of magnitude across backbones and
environments, the controller operates on a normalized cost
$\hat{c}_i \in [0,1]$: we maintain a running maximum
$C^{\max}_i = \max_{j \le i} c_j$ and set
$\hat{c}_i = c_i / C^{\max}_i$.  A cost weight $\lambda \ge 0$,
introduced in Eq.~(\ref{eq:select}), controls how strongly the
controller trades accuracy against cost: $\lambda{=}0$ recovers
pure accuracy maximization, while large $\lambda$ biases selection
toward the cheapest arm.  We use $\lambda{=}0.5$ in all reported
experiments.

Even when memory is relevant, injection comes at a cost: it lengthens the prompt, raises API spend, and can destabilize the agent's
action formatting. Our goal is therefore not merely to store memory, but to allocate it adaptively.

\subsection{Memory Actions}

Before solving task $T_i$, the controller chooses an action
$a_i \in \mathcal{A}$ specifying both a compute budget and a memory mode:
\begin{equation}
  m_i \in \{\texttt{none},\; \texttt{full},\; \texttt{compressed},\;
  \texttt{retrieved}\}.
\end{equation}

\textit{No replay} (\texttt{none}): the task is solved without injecting past
trajectories.

\textit{Full replay} (\texttt{full}): the prompt receives a fixed
number of recent successful trajectories
verbatim~\cite{brown2020fewshot}, corresponding to the standard
replay baseline in LifelongAgentBench.

\textit{Compressed replay} (\texttt{compressed}): past trajectories are
compressed using a LongLLMLingua style question aware wrapper before
injection.  The current task instruction serves as the compression query and
each replay trajectory is a separate context block.

\textit{Retrieved replay} (\texttt{retrieved}): the controller
retrieves the most relevant trajectories from a larger memory
bank, similar in spirit to retrieval-augmented
generation~\cite{lewis2020rag} but applied to prior interaction
traces rather than external documents.  In the DBBench
implementation, sessions are scored by
\begin{equation}
  r(\sigma_j, T_i)
  = w_{\ell}\,\mathrm{Jac}(q_i, q_j)
  + w_{s}\,\mathrm{Jac}(S_i, S_j)
  + w_{b}\,\mathbf{1}[b_i{=}b_j]
  + \epsilon,
\label{eq:retrieval}
\end{equation}
where $q_i$ is the tokenized instruction, $S_i$ is the skill set, $b_i$ is
the table identifier, and $\epsilon$ is a recency tie breaking bonus.  
The weights $(w_\ell, w_s, w_b)$ were selected through a small
hyperparameter-tuning pass on a held-out development subset, separate from all reported test runs, and held fixed across all retrieval-based variants (retrieved replay, retrieval$+$LongLLMLingua, and failure-aware
retrieval) so differences between methods are not attributable to
per-method retrieval retuning. The top $k$ retrieved trajectories are then injected. 

\paragraph{Trajectory rendering}
Trajectories from $\mathcal{M}_i$ can be injected verbatim (\textit{raw})
or in a deterministic \textit{trimmed-trace} rendering that compresses
verbose tool outputs while preserving the agent's action format.  Trimmed
traces truncate long environment responses (e.g., \texttt{stdout} blocks
in OS Interaction or table dumps in DBBench) and collapse repeated outputs,
but keep all agent-issued commands intact.  This contrasts with token-level
compression (e.g., LongLLMLingua), which can damage the structured action
syntax that interactive environments require.  Both raw and trimmed-trace
rendering are evaluated as baselines in Section~\ref{sec:experiments}.

\subsection{\Design{} Controller}

Before specifying the controller, we state four properties that
an inference-time memory allocator for lifelong agents must
satisfy.  \textbf{D1: Black-box compatibility.}  The controller
cannot require model weights, gradients, or internal
activations, so the same mechanism can allocate memory for both
open-weight and commercial API backbones.  \textbf{D2:
Single-pass online learning.}  The controller must adapt during
the task stream itself, without a separate offline training
phase or a warm-start dataset.  \textbf{D3: Bounded per-task
overhead.}  Controller decision cost must be small compared to
the LLM call it schedules; overhead that grows with the task
count is not viable at $500$-task horizons.  \textbf{D4:
Explicit cost objective.}  Memory injection consumes a shared
inference budget with retrieval, reasoning, and tool use, so
accuracy and cost must be traded as first-class objectives.  The
remainder of this section instantiates a controller that
satisfies all four.

We frame allocation as a \emph{contextual bandit} rather than full
reinforcement learning both to satisfy \textbf{D1} and \textbf{D2}
(weight-free updates driven by per-task outcomes), and because
per-task action selection has no long-horizon credit assignment
(the reward signal is observed before the next task arrives),
and rather than supervised learning because no ground-truth
optimal action is available for any task.  Contextual bandits~\cite{li2010linucb,auer2002ucb} are the
matching abstraction: they explore actions, observe outcomes, and
update online from a single scalar reward signal per decision. \Design{} is a gradient-free contextual bandit for inference-time
allocation, extending the LinUCB
formulation~\cite{li2010linucb} with a two-head parameterization
for joint accuracy and cost prediction.  It selects a budget primitive $a_i \in \mathcal{A}$ before each
task.  Each primitive specifies a replay mode, token budget, maximum round
count, and tool call budget.  The retrieval gated action set used in our DB
experiments is shown in Table~\ref{tab:primitives}.

\begin{table}[t]
\centering
\caption{Retrieval gated budget primitives used in DBBench experiments.}
\label{tab:primitives}
\begin{tabular}{lcccc}
\toprule
Primitive & Replay & Tokens & Rounds & Tools \\
\midrule
\texttt{no-replay-256}   & none      & 256 & 2 & 4 \\
\texttt{no-replay-512}   & none      & 512 & 3 & 6 \\
\texttt{full-512}        & full      & 512 & 3 & 6 \\
\texttt{retrieved-256}   & retrieved & 256 & 2 & 4 \\
\texttt{retrieved-512}   & retrieved & 512 & 3 & 6 \\
\texttt{retrieved-768}   & retrieved & 768 & 3 & 6 \\
\bottomrule
\end{tabular}
\end{table}

\noindent Compression gated experiments use analogous primitives with
\texttt{compressed} in place of \texttt{retrieved}.

\paragraph{Feature Vector}
Before each task $T_i$, the controller builds a feature vector
$\mathbf{x}_i \in \mathbb{R}^d$ from lightweight online signals:
\begin{multline}
  \mathbf{x}_i = \bigl[1,\; \tilde{p}_i,\; \tilde{u}_i,\; \tilde{s}_i,\;
  \tilde{f}_i,\; u_i/p_i,\; \tilde{\ell}_i,\; \\
  \bar{c}_i,\; \bar{r}_i,\; \hat{c}_{i-1},\;
  \mathbf{1}[y_{i-1}{=}1],\; \mathbf{1}[\mathrm{err}_{i-1}]
  \bigr]^{\!\top},
\label{eq:features}
\end{multline}
The $12$ features fall into three groups.
\textit{Prompt-shape features} ($\tilde{p}_i, \tilde{u}_i,
\tilde{s}_i, \tilde{f}_i, u_i/p_i, \tilde{\ell}_i$) are
log-normalized token counts for the full prompt, last user turn,
system prompt, and scaffold, together with the last-user share
and the normalized skill count.  These capture the size and
composition of the incoming task, distinguishing retrieval-heavy
prompts from tool-heavy ones and flagging tasks that would
exceed context if replay were added.
\textit{Runtime-state features} ($\bar{c}_i, \bar{r}_i,
\hat{c}_{i-1}$) are the EMA of normalized cost, the running
accuracy, and the previous sample's normalized cost.  These let
the controller condition on drift, shifting allocation when
recent tasks have been unusually expensive or unusually
successful without waiting for per-arm statistics to converge.
\textit{Previous-outcome features}
($\mathbf{1}[y_{i-1}{=}1], \mathbf{1}[\mathrm{err}_{i-1}]$) are
binary indicators for whether the last task succeeded and
whether it terminated in an error state; recent errors typically
indicate a budget or format failure the controller can respond
to by increasing the compute-budget tier.  All features are
computed before the LLM call for task $T_i$ and require no
weight access, satisfying \textbf{D1}; extraction is
$\mathcal{O}(|T_i|)$ in the prompt length, dominated by
tokenization that is required anyway.

\paragraph{Two Head LinUCB}
\Design{} maintains one precision matrix $A_a \in \mathbb{R}^{d \times d}$ per
action and two parameter vectors $\mathbf{b}_a^{\mathrm{acc}}$,
$\mathbf{b}_a^{\mathrm{cost}} \in \mathbb{R}^d$:
\begin{align}
  \hat{\theta}_a^{\mathrm{acc}}  &= A_a^{-1}\mathbf{b}_a^{\mathrm{acc}}, &
  \hat{\theta}_a^{\mathrm{cost}} &= A_a^{-1}\mathbf{b}_a^{\mathrm{cost}}.
\end{align}
Head 1 estimates the correctness probability,
$\hat{p}_a(\mathbf{x}) = \sigma(\hat{\theta}_a^{\mathrm{acc}\top}\mathbf{x})$,
and Head 2 estimates normalized expected cost,
$\hat{c}_a(\mathbf{x}) = \max(\hat{\theta}_a^{\mathrm{cost}\top}\mathbf{x}, 0)$.

\paragraph{Action Selection}
Given a cost weight $\lambda > 0$ and an exploration coefficient
$\alpha > 0$, the controller selects
\begin{equation}
  a_i^* = \arg\max_{a \in \mathcal{A}}\;
  \bigl[\hat{p}_a(\mathbf{x}_i) - \lambda\hat{c}_a(\mathbf{x}_i)\bigr]
  + \alpha\sqrt{\mathbf{x}_i^{\top} A_a^{-1}\mathbf{x}_i}.
\label{eq:select}
\end{equation}
Selection is performed before replay injection so that the chosen mode
determines how the prompt is constructed, making memory use a primary
decision variable rather than a fixed preprocessing step.

\begin{table*}[!tbp]
\centering
\caption{Overall cost--accuracy summary across three LLM backbones
and two environments on LifelongAgentBench (500 tasks each).  The
strongest fixed-memory baseline is the highest-accuracy fixed-memory
method available in each setting.  Cost reduction is \Design{}'s
cost relative to that baseline.  All \Design{} points use the
configuration described in Section~\ref{sec:setup}; per-cell method
points appear in Figure~\ref{fig:pareto-master}.  All values are
three-seed means (seeds $42$, $43$, $44$).}
\label{tab:summary}
\small
\begin{tabular}{llrrrr}
\toprule
Model & Env & No replay & Strongest baseline & \Design{} & Cost \\
      &     & (Acc./Cost) & (Acc./Cost)      & (Acc./Cost) & reduction \\
\midrule
Qwen 2.5 7B    & DB & 73.0 / \$0.046 & Fixed $k{=}16$: 74.85 / \$0.290       & 74.4 / \$0.050 & \textbf{$-82.8\%$} \\
Qwen 2.5 7B    & OS & 46.4 / \$0.104 & Fixed trim $k{=}1$: 51.5 / \$0.343   & 51.4 / \$0.226 & \textbf{$-34.1\%$} \\
Llama 3.1 8B   & DB & 20.8 / \$0.071 & Fixed $k{=}16$: 70.0 / \$0.316       & 66.2 / \$0.144 & \textbf{$-54.4\%$} \\
Llama 3.1 8B   & OS & 40.6 / \$0.234 & Fixed $k{=}1$: 44.4 / \$0.304        & 43.0 / \$0.224 & \textbf{$-26.3\%$} \\
GPT-4o mini    & DB & 66.4 / \$0.283 & LongLLMLingua: 69.0 / \$1.654        & 67.4 / \$0.292 & \textbf{$-82.3\%$} \\
GPT-4o mini    & OS & 59.0 / \$0.783 & LongLLMLingua: 61.8 / \$1.400        & 62.4 / \$0.833 & \textbf{$-40.5\%$} \\
\bottomrule
\end{tabular}
\end{table*}

\paragraph{Online Update}
After task $T_i$ completes, the bandit state is updated:
\begin{align}
  A_{a_i} &\leftarrow A_{a_i} + \mathbf{x}_i\mathbf{x}_i^{\top}, \\
  \mathbf{b}_{a_i}^{\mathrm{acc}}  &\leftarrow \mathbf{b}_{a_i}^{\mathrm{acc}}
    + y_i\mathbf{x}_i, \quad
  \mathbf{b}_{a_i}^{\mathrm{cost}} \leftarrow \mathbf{b}_{a_i}^{\mathrm{cost}}
    + \hat{c}_i\mathbf{x}_i.
\end{align}
The inverse $A_{a_i}^{-1}$ is maintained via the Sherman Morrison rank one
update in $\mathcal{O}(d^2)$:
\begin{equation}
  A_{a_i}^{-1} \leftarrow A_{a_i}^{-1}
  - \frac{(A_{a_i}^{-1}\mathbf{x}_i)(A_{a_i}^{-1}\mathbf{x}_i)^{\top}}
         {1 + \mathbf{x}_i^{\top} A_{a_i}^{-1}\mathbf{x}_i}.
\label{eq:sm}
\end{equation}
No gradients are computed; no model weights are modified.

\begin{algorithm}[t]
\caption{\Design{}: Two Head LinUCB Memory and Budget Allocation}
\label{alg:limbo}
\begin{algorithmic}[1]
\REQUIRE task stream $\mathcal{T}=(T_1,\ldots,T_N)$, action set
$\mathcal{A}$, dim $d$, exploration $\alpha$, cost weight $\lambda$
\ENSURE action sequence $(a_1^*,\ldots,a_N^*)$
\FOR{$a \in \mathcal{A}$}
  \STATE $A_a \leftarrow I_d$;\;
         $\mathbf{b}_a^{\mathrm{acc}} \leftarrow \mathbf{0}$;\;
         $\mathbf{b}_a^{\mathrm{cost}} \leftarrow \mathbf{0}$
\ENDFOR
\STATE $\bar{c} \leftarrow 0$;\; $\bar{r} \leftarrow 0$
\FOR{$i = 1$ \textbf{to} $N$}
  \STATE Build $\mathbf{x}_i$ from $T_i$ and runtime state
  \FOR{$a \in \mathcal{A}$}
    \STATE $\hat{p}_a \leftarrow \sigma(A_a^{-1}\mathbf{b}_a^{\mathrm{acc}}\cdot\mathbf{x}_i)$;\;
           $\hat{c}_a \leftarrow \max(A_a^{-1}\mathbf{b}_a^{\mathrm{cost}}\cdot\mathbf{x}_i,\,0)$
    \STATE $\mathrm{score}(a) \leftarrow \hat{p}_a - \lambda\hat{c}_a
           + \alpha\sqrt{\mathbf{x}_i^{\top}A_a^{-1}\mathbf{x}_i}$
  \ENDFOR
  \STATE $a_i^* \leftarrow \arg\max_{a}\,\mathrm{score}(a)$
  \STATE Set replay mode from $a_i^*$; construct prompt accordingly
  \STATE Execute $T_i$; observe $y_i\in\{0,1\}$ and cost $c_i$
  \STATE Sherman Morrison update $A_{a_i^*}^{-1}$ (Eq.~\ref{eq:sm})
  \STATE $\mathbf{b}_{a_i^*}^{\mathrm{acc}}\mathrel{+}=y_i\mathbf{x}_i$;\;
         $\mathbf{b}_{a_i^*}^{\mathrm{cost}}\mathrel{+}=\hat{c}_i\mathbf{x}_i$
  \STATE Update EMA $\bar{c}$ and running accuracy $\bar{r}$
\ENDFOR
\end{algorithmic}
\end{algorithm}

\paragraph{Complexity}
Per task, \Design{} performs one Sherman--Morrison update
(Eq.~\ref{eq:sm}) at $\mathcal{O}(d^2)$ per action and
$|\mathcal{A}|$ dot products to score arms, for
$\mathcal{O}(|\mathcal{A}|\,d^2)$ total, with $d=12$ and
$|\mathcal{A}| \le 12$ in our experiments.  Controller overhead
is under $0.5$~ms per task on a single CPU core, negligible next
to the seconds of LLM inference it schedules.  Memory footprint
per action is $\mathcal{O}(d^2)$, dominated by the precision
matrix $A_a$: $12 \times 12 \times 8$ bytes $\approx 1.2$~KB per
arm.  \Design{} therefore satisfies \textbf{D3} with substantial
headroom, and scales linearly in $|\mathcal{A}|$ if the action
space is later enlarged.

\paragraph{Full Algorithm and Online Adaptation}
Algorithm~\ref{alg:limbo} summarizes the complete online loop. On each task, \Design{} interleaves action selection (lines~7--12) with the post-task bandit update (lines~15--17). Since there is no separate offline training phase, the controller continuously adapts throughout deployment.

\subsection{Instantiation in This Work}

We study \Design{} through three action space designs:
\begin{enumerate}
  \item \textbf{Compression gated}: arms choose among no replay, full recent
        replay, and LongLLMLingua compressed replay.
  \item \textbf{Retrieval gated}: arms choose among no replay, full recent
        replay, and relevance retrieved replay (top $k$ from a 64 session
        memory bank, with $k \in \{4, 8, 16\}$ tested as separate configs).
  \item \textbf{Retrieval gated plus rescue}: the retrieval gated bandit is
        augmented with a confidence triggered second pass that reruns
        low confidence samples with maximum retrieved context.
\end{enumerate}
This progression separates which memories are worth keeping from
when any memory is worth spending tokens on.  \Design{} addresses the
second question directly while remaining compatible with multiple answers to
the first.
 \section{Experimental Analysis}
\label{sec:experiments}

\subsection{Setup}
\label{sec:setup}
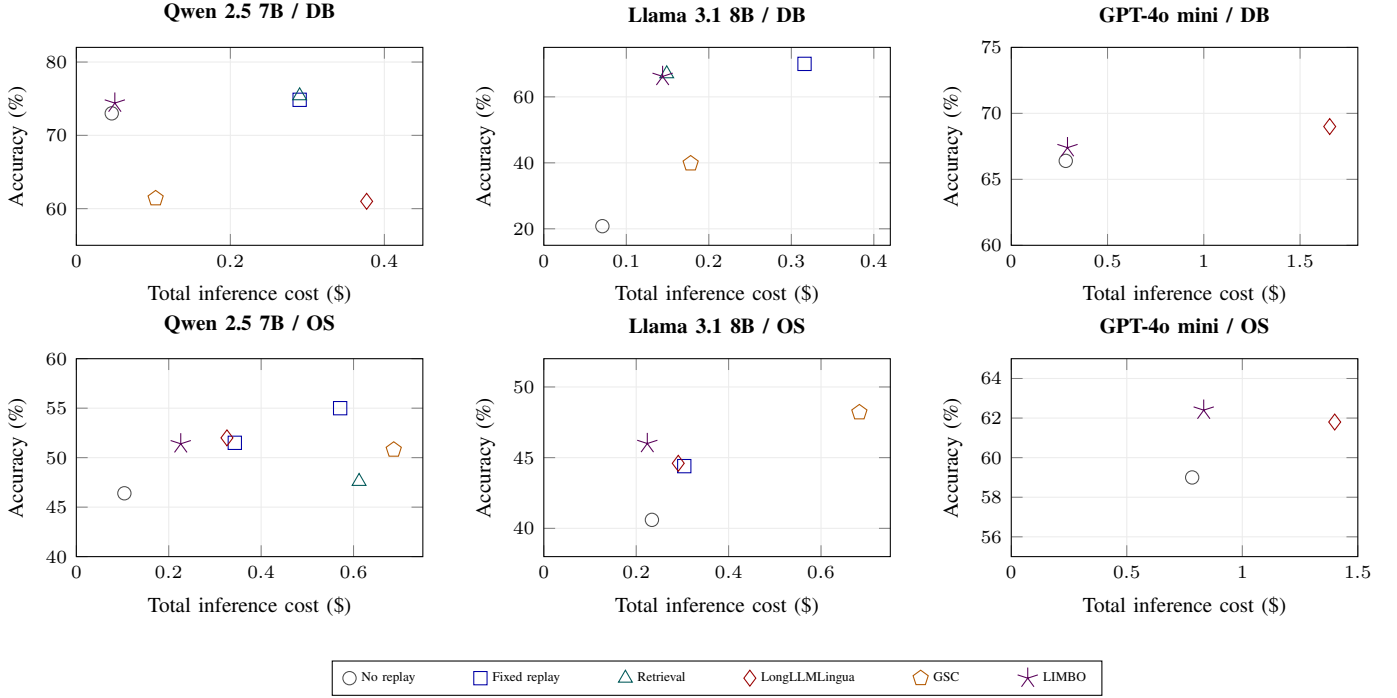
\begin{figure*}[t]
\centering
\begin{tikzpicture}

\begin{groupplot}[
    group style={
        group size=3 by 2,
        horizontal sep=1.6cm,
        vertical sep=1.5cm,
    },
    width=0.34\textwidth,
    height=4.2cm,
    xlabel={Total inference cost (\$)},
    ylabel={Accuracy (\%)},
    grid=major,
    grid style={gray!15},
    tick label style={font=\scriptsize},
    label style={font=\footnotesize},
    title style={font=\footnotesize\bfseries},
]


\nextgroupplot[
    title={Qwen 2.5 7B / DB},
    xmin=0, xmax=0.45,
    ymin=55, ymax=82
]

\addplot[
    only marks,
    mark=o,
    mark size=2.5pt,
    color=gray!60!black
] coordinates {(0.046,73.0)};

\addplot[
    only marks,
    mark=square,
    mark size=2.5pt,
    color=blue!65!black
] coordinates {(0.290,74.85)};

\addplot[
    only marks,
    mark=triangle,
    mark size=3pt,
    color=teal!70!black
] coordinates {(0.290,75.4)};

\addplot[
    only marks,
    mark=diamond,
    mark size=3pt,
    color=red!60!black
] coordinates {(0.377,61.0)};

\addplot[
    only marks,
    mark=pentagon,
    mark size=3pt,
    color=orange!70!black
] coordinates {(0.103,61.4)};

\addplot[
    only marks,
    mark=star,
    mark size=4pt,
    color=violet!70!black
] coordinates {(0.050,74.4)};


\nextgroupplot[
    title={Llama 3.1 8B / DB},
    xmin=0, xmax=0.42,
    ymin=15, ymax=75
]

\addplot[
    only marks,
    mark=o,
    mark size=2.5pt,
    color=gray!60!black
] coordinates {(0.071,20.8)};

\addplot[
    only marks,
    mark=square,
    mark size=2.5pt,
    color=blue!65!black
] coordinates {(0.316,70.0)};

\addplot[
    only marks,
    mark=triangle,
    mark size=3pt,
    color=teal!70!black
] coordinates {(0.149,67.0)};

\addplot[
    only marks,
    mark=pentagon,
    mark size=3pt,
    color=orange!70!black
] coordinates {(0.178,39.8)};

\addplot[
    only marks,
    mark=star,
    mark size=4pt,
    color=violet!70!black
] coordinates {(0.144,66.2)};


\nextgroupplot[
    title={GPT-4o mini / DB},
    xmin=0, xmax=1.8,
    ymin=60, ymax=75
]

\addplot[
    only marks,
    mark=o,
    mark size=2.5pt,
    color=gray!60!black
] coordinates {(0.283,66.4)};

\addplot[
    only marks,
    mark=diamond,
    mark size=3pt,
    color=red!60!black
] coordinates {(1.654,69.0)};

\addplot[
    only marks,
    mark=star,
    mark size=4pt,
    color=violet!70!black
] coordinates {(0.292,67.4)};


\nextgroupplot[
    title={Qwen 2.5 7B / OS},
    xmin=0, xmax=0.75,
    ymin=40, ymax=60
]

\addplot[
    only marks,
    mark=o,
    mark size=2.5pt,
    color=gray!60!black
] coordinates {(0.104,46.4)};

\addplot[
    only marks,
    mark=square,
    mark size=2.5pt,
    color=blue!65!black
] coordinates {
    (0.343,51.5)
    (0.571,55.0)
};

\addplot[
    only marks,
    mark=triangle,
    mark size=3pt,
    color=teal!70!black
] coordinates {(0.612,47.6)};

\addplot[
    only marks,
    mark=diamond,
    mark size=3pt,
    color=red!60!black
] coordinates {(0.326,52.0)};

\addplot[
    only marks,
    mark=pentagon,
    mark size=3pt,
    color=orange!70!black
] coordinates {(0.687,50.8)};

\addplot[
    only marks,
    mark=star,
    mark size=4pt,
    color=violet!70!black
] coordinates {(0.226,51.4)};


\nextgroupplot[
    title={Llama 3.1 8B / OS},
    xmin=0, xmax=0.75,
    ymin=38, ymax=52
]

\addplot[
    only marks,
    mark=o,
    mark size=2.5pt,
    color=gray!60!black
] coordinates {(0.234,40.6)};

\addplot[
    only marks,
    mark=square,
    mark size=2.5pt,
    color=blue!65!black
] coordinates {(0.304,44.4)};

\addplot[
    only marks,
    mark=diamond,
    mark size=3pt,
    color=red!60!black
] coordinates {(0.291,44.6)};

\addplot[
    only marks,
    mark=pentagon,
    mark size=3pt,
    color=orange!70!black
] coordinates {(0.683,48.2)};

\addplot[
    only marks,
    mark=star,
    mark size=4pt,
    color=violet!70!black
] coordinates {(0.224,46.0)};

\nextgroupplot[
    title={GPT-4o mini / OS},
    xmin=0, xmax=1.5,
    ymin=55, ymax=65,
    legend to name=paretoleg,
    legend columns=6,
    legend style={
        font=\tiny,
        draw=black,
        fill=white,
        at={(0.5,-0.35)},
        anchor=north,
        /tikz/every even column/.append style={column sep=0.7cm},
        inner xsep=4pt,
        inner ysep=2pt,
    }
]

\addplot[
    only marks,
    mark=o,
    mark size=2.5pt,
    color=gray!60!black,
    forget plot
] coordinates {(0.783,59.0)};

\addplot[
    only marks,
    mark=diamond,
    mark size=3pt,
    color=red!60!black,
    forget plot
] coordinates {(1.400,61.8)};

\addplot[
    only marks,
    mark=star,
    mark size=4pt,
    color=violet!70!black,
    forget plot
] coordinates {(0.833,62.4)};


\addlegendimage{
    only marks,
    mark=o,
    mark size=2.5pt,
    color=gray!60!black
}
\addlegendentry{No replay}

\addlegendimage{
    only marks,
    mark=square,
    mark size=2.5pt,
    color=blue!65!black
}
\addlegendentry{Fixed replay}

\addlegendimage{
    only marks,
    mark=triangle,
    mark size=3pt,
    color=teal!70!black
}
\addlegendentry{Retrieval}

\addlegendimage{
    only marks,
    mark=diamond,
    mark size=3pt,
    color=red!60!black
}
\addlegendentry{LongLLMLingua}

\addlegendimage{
    only marks,
    mark=pentagon,
    mark size=3pt,
    color=orange!70!black
}
\addlegendentry{GSC}

\addlegendimage{
    only marks,
    mark=star,
    mark size=4pt,
    color=violet!70!black
}
\addlegendentry{LIMBO}

\end{groupplot}

\node[anchor=north]
at ($(group c2r2.south) - (0,1.25cm)$)
{\ref{paretoleg}};

\end{tikzpicture}

\caption{
Cost--accuracy plots across three LLM backbones (columns) and two
environments (rows). Markers denote no replay (circles), fixed replay
(squares), retrieval (triangles), LongLLMLingua (diamonds), GSC
(pentagons), and LIMBO (stars). LIMBO generally occupies a low-cost,
competitive-accuracy region, while the highest-accuracy fixed-replay
configurations often incur substantially greater inference cost.
}

\label{fig:pareto-master}
\end{figure*}
\paragraph{Benchmark and tasks}
We evaluate on LifelongAgentBench (LAB)~\cite{zheng2025lab} using DB
(SQL) and OS (bash) environments, each with the 500-task sequential
slice released with LAB.  Three backbones---Qwen2.5 7B Instruct,
Llama 3.1 8B Instruct, and GPT-4o mini---are evaluated on both
environments, yielding six (model, environment) settings.  All
reported \Design{} and baseline points are averaged over three
matched seeds ($42$, $43$, $44$); within-cell comparisons use
matched seeds throughout.

\paragraph{Baselines}
Throughout, $k$ denotes the number of past successful trajectories
injected per task.  We compare against:
\textbf{No replay} (zero-shot per task);
\textbf{Fixed replay ($k{=}N$)}, the $N$ most recent successful
trajectories injected verbatim in a deterministic trimmed-trace
rendering that preserves executable action structure (this
consistently beats token-level compression at matched budget, e.g.,
$51.5\%$ vs.\ $49.4\%$ on Qwen OS at $k{=}1$);
\textbf{LongLLMLingua}~\cite{jiang2024longllmlingua}, question-aware
token-level compression applied to the same recent-replay buffer;
and \textbf{Group Self-Consistency (GSC)} at $\text{usc}{=}1$,
LAB's native voting-based stabilizer~\cite{zheng2025lab}.

\paragraph{The \Design{} configuration used across cells}
All six reported \Design{} points use the same controller
formulation, the same feature vector (Eq.~(\ref{eq:features})),
the same online update, and the same 4-way memory action space
(\texttt{none}, \texttt{full}, \texttt{compressed},
\texttt{retrieved}) paired with three generation-budget tiers (low
$\leq 512$, medium $768$, high $\geq 1024$).  Retrieval scoring
weights in Eq.~(\ref{eq:retrieval}) are $w_{\ell}{=}0.35$,
$w_{s}{=}0.50$, $w_{b}{=}0.15$ and are held fixed across every
retrieval-based method.  The memory-bank size is $64$ in every cell.
The environment-specific choice is the retrieved-$k$ selected inside
the \texttt{retrieved} action: DB cells expose top-$8$, OS cells
expose top-$1$, matching the strongest fixed-replay setting for each
environment.

\paragraph{Metrics and infrastructure}
We report accuracy and total token cost (USD-equivalent per 500-task
run).  Open-weight backbones run locally through Hugging Face
\texttt{transformers} in \texttt{bfloat16} on NVIDIA RTX PRO 6000
GPUs; GPT-4o mini via the OpenAI API.  Decoding is greedy with a
512-token per-response cap.  Costs use provider API rates for
GPT-4o mini and a standardized \$0.04/\$0.10 per million
input/output tokens for local 7B/8B models; local-model dollar
values are standardized proxies rather than measured serving
expense.

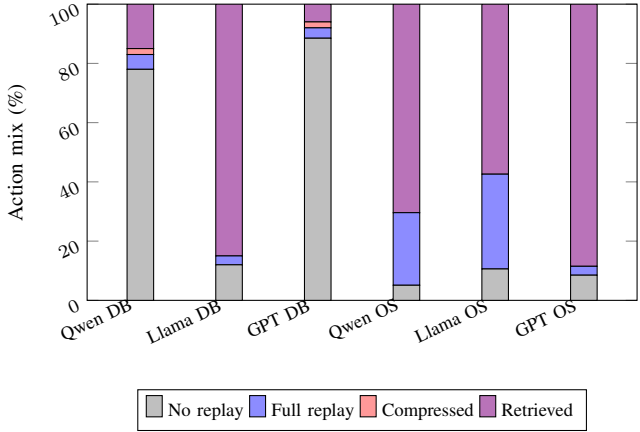
\begin{figure}[t]
\centering
\begin{tikzpicture}
\begin{axis}[
    ybar stacked,
    width=\columnwidth, height=5.5cm,
    ylabel={Action mix (\%)},
    symbolic x coords={Qwen DB, Llama DB, GPT DB, Qwen OS, Llama OS, GPT OS},
    xtick=data,
    ymin=0, ymax=100,
    legend style={font=\scriptsize, at={(0.5,-0.3)}, anchor=north, legend columns=4},
    tick label style={font=\scriptsize, rotate=25, anchor=east},
    label style={font=\footnotesize},
    enlarge x limits=0.12,
]
\addplot[fill=gray!50] coordinates
  {(Qwen DB, 78.0) (Llama DB,  12.0) (GPT DB, 88.5) (Qwen OS,  5.1) (Llama OS, 10.6) (GPT OS,  8.5)};
\addplot[fill=blue!45] coordinates
  {(Qwen DB,  5.0) (Llama DB,  3.0) (GPT DB,  3.5) (Qwen OS, 24.5) (Llama OS, 32.0) (GPT OS,  3.0)};
\addplot[fill=red!40] coordinates
  {(Qwen DB,  2.0) (Llama DB,  0.0) (GPT DB,  2.0) (Qwen OS,  0.0) (Llama OS,  0.0) (GPT OS,  0.0)};
\addplot[fill=violet!55] coordinates
  {(Qwen DB,  15.0) (Llama DB, 85.0) (GPT DB,  6.0) (Qwen OS, 70.4) (Llama OS, 57.4) (GPT OS, 88.5)};
\legend{No replay, Full replay, Compressed, Retrieved}
\end{axis}
\end{tikzpicture}
\caption{\Design{}'s learned action mix across the six
(model, environment) settings, each a three-seed mean
(seeds $42$, $43$, $44$).  The same controller, features, and hyperparameters produce sharply different policies per cell.  Qwen DB and GPT-4o mini DB stay near no-replay ($\approx 78\%$ of
tasks); Llama DB shifts $85\%$ of tasks to retrieved replay to
recover from a $20.8\%$ no-replay collapse; all three OS
environments concentrate on retrieved and full replay, with mix
proportions that vary by backbone.  No fixed replay policy is
uniformly cost-effective, and each cell retains task-level
variation rather than collapsing to a single arm.}
\label{fig:action-mix}
\end{figure}
\subsection{Overall Cost--Accuracy Tradeoff}
\label{sec:overall}

Table~\ref{tab:summary} and Figure~\ref{fig:pareto-master} summarize
\Design{}'s performance across all six (model, environment) settings.

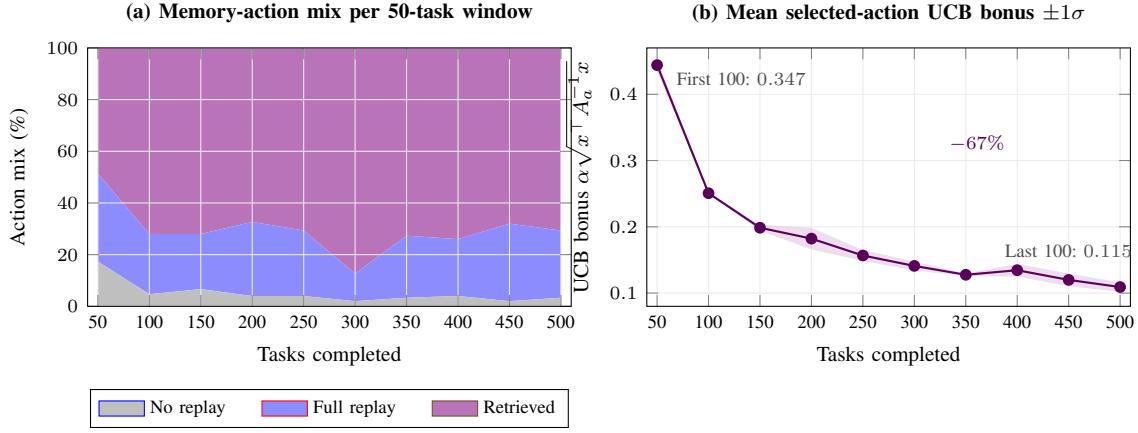
\begin{figure*}[!tbp]
\centering
\begin{tikzpicture}
\begin{groupplot}[
    group style={
        group size=2 by 1,
        horizontal sep=1.0cm,
    },
    width=0.44\textwidth, height=5.0cm,
    xmin=0.8, xmax=10.2,
    xtick={1,2,3,4,5,6,7,8,9,10},
    xticklabels={50,100,150,200,250,300,350,400,450,500},
    xlabel={Tasks completed},
    tick label style={font=\scriptsize},
    label style={font=\footnotesize},
    title style={font=\footnotesize\bfseries},
    grid=major, grid style={gray!15},
]

\nextgroupplot[
    ymin=0, ymax=100,
    ylabel={Action mix (\%)},
    title={(a) Memory-action mix per 50-task window},
    stack plots=y,
    area style,
    legend style={font=\scriptsize, at={(0.5,-0.32)}, anchor=north, legend columns=3, /tikz/every even column/.append style={column sep=0.4cm}},
]
\addplot+[mark=none, draw=none, fill=gray!50] coordinates {
  (1,17.333) (2,4.667) (3,6.667) (4,4.000) (5,4.000)
  (6,2.000)  (7,3.333) (8,4.000) (9,2.000) (10,3.333)
} \closedcycle;
\addlegendentry{No replay}

\addplot+[mark=none, draw=none, fill=blue!45] coordinates {
  (1,34.000) (2,23.333) (3,21.333) (4,28.667) (5,25.333)
  (6,10.667) (7,24.000) (8,22.000) (9,30.000) (10,26.000)
} \closedcycle;
\addlegendentry{Full replay}

\addplot+[mark=none, draw=none, fill=violet!55] coordinates {
  (1,48.667) (2,72.000) (3,72.000) (4,67.333) (5,70.667)
  (6,87.333) (7,72.667) (8,74.000) (9,68.000) (10,70.667)
} \closedcycle;
\addlegendentry{Retrieved}

\nextgroupplot[
    ymin=0.08, ymax=0.47,
    ylabel={UCB bonus $\alpha\sqrt{x^{\top} A_a^{-1} x}$},
    title={(b) Mean selected-action UCB bonus $\pm 1\sigma$},
    scaled y ticks=false,
    ytick={0.1,0.2,0.3,0.4},
]
\addplot[draw=none, name path=upper] coordinates {
    (1,0.4537) (2,0.2515) (3,0.2030) (4,0.1985) (5,0.1649)
    (6,0.1477) (7,0.1294) (8,0.1441) (9,0.1294) (10,0.1163)};
\addplot[draw=none, name path=lower] coordinates {
    (1,0.4345) (2,0.2500) (3,0.1942) (4,0.1659) (5,0.1486)
    (6,0.1342) (7,0.1258) (8,0.1250) (9,0.1104) (10,0.1020)};
\addplot[fill=violet!30, draw=none, opacity=0.45]
    fill between[of=upper and lower];

\addplot[
    thick, color=violet!70!black,
    mark=*, mark size=1.8pt,
] coordinates {
    (1,0.4441) (2,0.2507) (3,0.1986) (4,0.1822) (5,0.1567)
    (6,0.1410) (7,0.1276) (8,0.1345) (9,0.1199) (10,0.1091)};

\node[font=\scriptsize, anchor=south west, text=black!70]
    at (axis cs:1.2,0.40) {First 100: $0.347$};
\node[font=\scriptsize, anchor=south, text=black!70]
    at (axis cs:9.0,0.14) {Last 100: $0.115$};
\node[font=\scriptsize\bfseries, anchor=south west, text=violet!70!black]
    at (axis cs:6.5,0.30) {$-67\%$};

\end{groupplot}
\end{tikzpicture}
\caption{\Design{}'s online adaptation over the 500-task Qwen OS
stream, averaged across three seeds ($42$, $43$, $44$) in
non-overlapping 50-task windows.  (a) The memory-action mix shifts
from an exploration-heavy early regime (window 1: $17.3\%$
no replay, $34.0\%$ full replay, $48.7\%$ retrieved) toward a
stable replay-heavy regime (windows 2--10 mean: $3.8\%$ no
replay, $23.5\%$ full replay, and $72.7\%$ retrieved replay).
Compressed replay is selected on $0\%$ of tasks throughout and is
omitted from the stack.
(b) The $\alpha$-scaled UCB uncertainty bonus for the selected
action decays by $67\%$ from the first 100 tasks to the last 100
($0.347 \to 0.115$) as per-action precision matrices $A_a$
accumulate observations.  Cross-seed SDs (shaded band) stay below
$0.017$ in every window, so the trajectory is a property of the
controller rather than of any single run.}
\label{fig:online-dynamics}
\end{figure*}

Two patterns hold across all six settings.  First, \Design{} lies on
the empirical cost--accuracy Pareto frontier in every cell: no
fixed method is simultaneously cheaper and more accurate.  Second,
across the six cells \Design{} matches or exceeds the strongest
memory-augmented baseline's accuracy (within $-3.8$ to $+0.6$
points) while reducing cost by $26$--$83\%$; on GPT-4o mini OS it
strictly Pareto-dominates the LongLLMLingua baseline, exceeding
its accuracy by $0.6$ points at $40.5\%$ lower cost.

The comparison against voting-based stabilization is clean.  LAB's
GSC baseline is strictly dominated by \Design{} in all four
matched settings (Qwen DB, Qwen OS, Llama DB, Llama OS): higher
accuracy \emph{and} lower cost.  On Llama DB the accuracy gap is
$+26$ points at lower cost.  Voting cannot recover replay value when
the underlying replay is applied indiscriminately.

The largest cost reductions ($\sim 83\%$) occur where the strongest
baseline commits to a $k{=}16$-scale replay budget (Qwen DB and
GPT-4o mini DB).  In these cells the no-replay baseline is already
strong, so raw accuracy differences are modest and the cost saving
carries the story.  The Llama DB cell tells the opposite story:
no-replay performs poorly at $20.8\%$, and \Design{} reaches
$66.2\%$ ($+45.4$ points over no-replay, within $3.8$ points of
fixed-replay) at less than half the fixed-replay cost.  This is the
clearest memory-essential regime in our evaluation.  On GPT-4o mini OS, \Design{} strictly Pareto-dominates the
LongLLMLingua baseline ($+0.6$ points at $40.5\%$ lower cost) and
lifts accuracy by $+3.4$ points over no-replay at essentially the
same cost.

\subsection{Policy Across Backbones and Environments}
\label{sec:regimes}

Figure~\ref{fig:action-mix} shows the memory-action mix \Design{}
converges to across all six settings under the same controller,
features, action space, and hyperparameters (retrieved-$k$ set as
described in Section~\ref{sec:setup}).  The learned policy differs
sharply across cells.

\paragraph{DB cells (Qwen, GPT-4o mini): low-replay regime}
No-replay accuracy is already high ($73.0\%$ and $66.4\%$
respectively) and injecting past trajectories yields only
$2$--$3$ points of accuracy at a $5$--$6\times$ cost multiplier.
\Design{} places $\approx 78\%$ of tasks on no-replay paired with
the lowest generation-budget tier, spending negligible additional
memory or compute.

\paragraph{Qwen OS: moderate-replay regime}
Replay meaningfully improves accuracy ($+5$ points from no-replay
to fixed $k{=}1$) but each additional trajectory adds substantial
cost.  \Design{} adopts a mixed policy: retrieved $k{=}1$ on
$\approx 70\%$ of tasks, full replay on $\approx 25\%$, with the
higher generation-budget tiers used more than half the time.  The
result matches fixed-replay accuracy at $34\%$ lower cost.

\paragraph{Llama DB: memory-essential regime}
Llama performs poorly on DB without prior demonstrations
($20.8\%$).  \Design{} shifts $\approx 85\%$ of tasks onto
retrieved replay while still keeping the majority of tasks in the
lowest generation-budget tier, recovering $66.2\%$ accuracy at
less than half the strongest baseline's cost.  The controller
invests in memory when the backbone requires it rather than
defaulting to a cheap policy.

\paragraph{GPT-4o mini OS: high-per-token-cost regime}
Under commercial API pricing, \Design{} selects retrieved $k{=}1$
on $\approx 88\%$ of tasks and pairs it with the medium
generation tier; because the retrieved arm carries a small
per-task memory footprint, aggregate cost remains within
$\sim 6\%$ of no-replay while accuracy improves by $+3.4$ points.
We describe this as an observed operating point rather than a
demonstrated causal adaptation to the pricing schedule; a direct
cost-structure ablation is future work.

The unifying observation is that the optimal
memory-and-compute policy is not a global constant.  It depends
on the backbone's raw capability, the environment's
replay-sensitivity, and the deployment's per-token cost, and
\Design{} exposes this variation online without any per-cell
tuning or offline retraining.

\subsection{Online Adaptation over the Task Stream}
\label{sec:online-dynamics}

Figure~\ref{fig:action-mix} shows the converged policy after the
full 500-task stream but does not by itself distinguish a bandit
that learns online from one that quickly settles into a fixed
mixture.  Figure~\ref{fig:online-dynamics} closes this gap on
Qwen OS, the setting with the most mixed converged policy.  Panel
(a) shows that the memory-action mix in the first 50 tasks is
substantially more exploratory than the converged mix: no-replay
selections drop from $17.3\%$ to a windows-2--10 mean of $3.8\%$,
full replay from $34.0\%$ to $23.5\%$, and retrieved replay rises
from $48.7\%$ to $72.7\%$.  Panel (b) provides internal-state
evidence for the same trajectory: the $\alpha$-scaled UCB
uncertainty bonus $\alpha\sqrt{x^{\top} A_a^{-1} x}$ for the
selected action decreases by $67\%$ from the first 100 tasks to
the last 100 ($0.347 \to 0.115$) because the per-action precision
matrices $A_a$ accumulate observations across the single-pass
stream.  The decline is monotonic in eight of nine window
transitions, with one small reversal at window 7 to 8 that stays
within the cross-seed standard deviation.  Together, the shift in
selected action mix and the contraction of the associated
exploration bonus provide direct evidence that the converged
policy in Figure~\ref{fig:action-mix} is learned during the stream
rather than fixed in advance.

\subsection{Discussion}
\label{sec:discussion}
Across three backbones and two environments, \Design{} occupies
the cost-efficient region of the Pareto frontier that fixed replay
and voting-based stabilization do not reach.  On Qwen OS it
matches the fixed-replay baseline within $0.1$ points at $34\%$
lower cost; on Llama DB it stays within $3.8$ points of fixed
$k{=}16$ at $54\%$ lower cost, in a regime where no-replay
collapses to $20.8\%$; on Qwen DB and GPT-4o mini DB it recognizes
that additional memory does not pay off and stays within $2$
accuracy points of the strongest baseline at roughly one-sixth the
cost; on GPT-4o mini OS it strictly Pareto-dominates the LongLLMLingua
baseline ($+0.6$ points at $40\%$ lower cost) while improving on
no-replay accuracy by $+3.4$ points.  In every cell the learned action mix
differs qualitatively across model and environment, and the same
controller instantiation reaches those different mixes online
without per-cell tuning or offline retraining.

DBBench plays the role of a control environment on strong
backbones, where the controller's task is largely to recognize
that additional memory does not pay off and stay cheap.
\Design{} does this successfully, matching the standard
fixed-replay baseline at one-sixth its cost.  \Design{}'s
contribution across settings is therefore \emph{frontier
competitiveness}: recovering most of the accuracy of the strongest
memory-augmented baseline while operating close to the no-replay
cost floor. 

\Design{} focuses on inference time memory allocation rather than memory representation. It acts as a cost aware control layer that decides when to use memory and how much compute to allocate, complementing methods that improve how memories are stored or retrieved. For example, while MemRL~\cite{zhang2026memrl} optimizes memory retrieval utility on strong backbones, \Design{} learns whether replay is beneficial for the current task and what budget should be allocated to it. Improved memory representations can be naturally integrated as additional replay actions in \Design{}'s action space.

\section{Conclusion}
\label{sec:conclusion}
We introduced \Design{}, the first online inference-time memory allocation framework for lifelong LLM agents. \Design{} uses a gradient-free contextual bandit to jointly select replay strategies and inference budgets without modifying the underlying agent. Across three LLM backbones on LifelongAgentBench, \Design{} achieves better cost--accuracy tradeoffs than state-of-the-art memory-augmented baselines, maintaining comparable performance while reducing inference cost by up to $\sim$83\% ($\sim$53\% on average). The learned policies adapt across models and environments, demonstrating that effective memory allocation can be discovered online rather than manually specified. These results highlight a broader opportunity for cost-aware online control of inference-time resources in lifelong LLM agents.

\section*{Acknowledgements}
This work has been funded in part by NSF, with award numbers \#2112665, \#2112167, \#2003279, \#2120019, \#2211386, \#2052809, \#1911095 and in part by PRISM and CoCoSys, centers in JUMP 2.0, an SRC program sponsored by DARPA.

\bibliographystyle{IEEEtran}
\bibliography{references}

\end{document}